\documentclass[letterpaper, 10 pt, conference]{ieeeconf}  

\IEEEoverridecommandlockouts                              

\usepackage{multicol}
\usepackage{tabularx}
\usepackage{caption}
\usepackage{graphicx}
\usepackage{amsfonts}
\usepackage{amsmath}
\usepackage{marvosym} 
\usepackage{bbm}
\makeatletter  
\newif\if@restonecol  
\makeatother

\usepackage[linesnumbered,ruled,vlined]{algorithm2e}
\usepackage{algpseudocode}

\title{\LARGE \bf
GraspARL: Dynamic Grasping via Adversarial Reinforcement Learning
}

\author{
Tianhao Wu, Fangwei Zhong, Yiran Geng, Hongchen Wang, Yongjian Zhu, Yizhou Wang, Hao Dong
\thanks{T. Wu, Y. Geng, H. Wang, Y. Zhu, Y. Wang, and H. Dong are with CFCS, School of CS, Peking University, Beijing, China.}%
\thanks{F. Zhong is with School of AI, Peking University, Beijing, China.}%
\thanks{T. Wu and F. Zhong are co-first authors.}
\thanks{Corresponding author: H. Dong (Email: hao.dong@pku.edu.cn).}%
}

\begin{document}

\maketitle
\thispagestyle{empty}
\pagestyle{empty}

\def\eg{\emph{e.g}.} \def\Eg{\emph{E.g}.}
\def\ie{\emph{i.e}.} \def\Ie{\emph{I.e}.}
\def\cf{\emph{c.f}.} \def\Cf{\emph{C.f}.}
\def\etc{\emph{etc}.} \def\vs{\emph{vs}.}
\def\wrt{w.r.t. } \def\dof{d.o.f. }
\def\etal{\emph{et al}. }

\begin{abstract}

Grasping moving objects, such as goods on a belt or living animals, is an important but challenging task in robotics.
Conventional approaches rely on a set of manually defined object motion patterns for training, resulting in poor generalization to unseen object trajectories.
In this work, we introduce an adversarial reinforcement learning framework for dynamic grasping, namely GraspARL. 
To be specific. we formulate the dynamic grasping problem as a “move-and-grasp” game, where the robot is to pick up the object on the mover and the adversarial mover is to find a path to escape it.
Hence, the two agents play a min-max game and are trained by reinforcement learning.
In this way, the mover can auto-generate diverse moving trajectories while training. And the robot trained with the adversarial trajectories can generalize to various motion patterns.
Empirical results on the simulator and real-world scenario demonstrate the effectiveness of each and good generalization of our method.
\end{abstract}

\section{INTRODUCTION}
In this paper, we study a practical yet challenging robot arm control task: moving object grasping / dynamic grasping~\cite{akinola2021dynamic,islam2021provably,migimatsu2020object}.
In this task, the agent should manipulate a robot arm to pick up a moving object in a high-dynamic environment. 
It is widely demanded in real-world applications, \emph{e.g.}, grasping objects on a conveyor belt, picking up a rolling ball, or picking up a bottle handed by a human.

Recent advances in deep learning have brought static object grasping to a new level in accuracy, robustness, and efficiency~\cite{morrison2018closing,kalashnikov2018scalable,mousavian20196,fang2020graspnet}. However, moving object grasping remains challenging. 
First, the perception module has to localize the moving object and model the dynamic in real-time, as the object will be occlude by others and the moving trajectories are noisy and of diverse patterns. 
Second, the agent should adaptively find a suitable grasping point according to the change of the state. 

\begin{figure}[t]
  \centering
  \includegraphics[trim=0 50 220 0,clip, width=\linewidth]{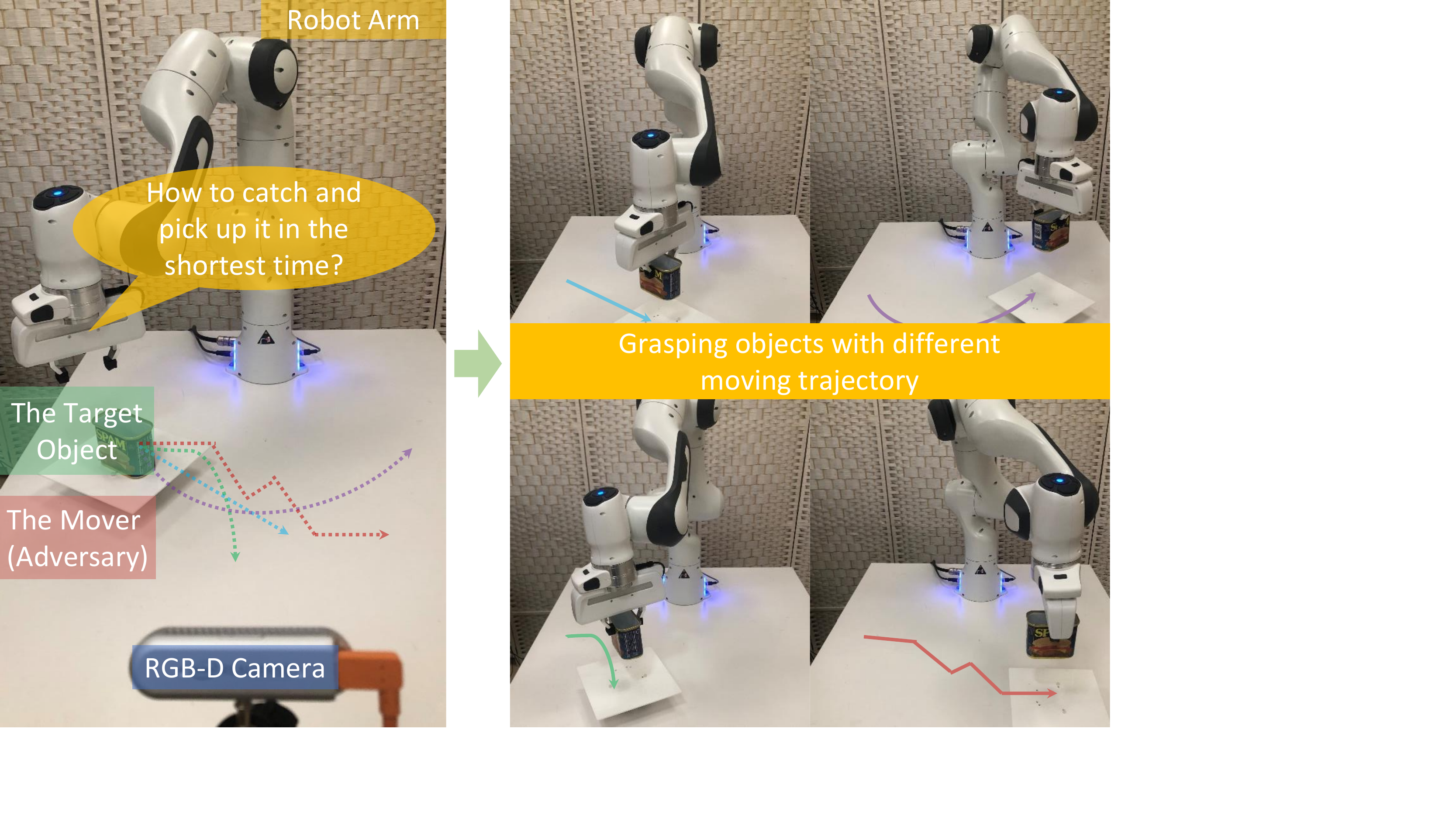}
  \caption{\textbf{Overview}. The robot arm and the mover play an adversarial game, the mover will learn to generate diverse trajectories (Left), which will improve the generalization of dynamic grasping for the robot arm (Right).}
  \vspace{-4mm}
  \label{fig:teaser}
\end{figure}

As is known to all, data is the key to building a robust deep learning network.
However, it is not easy to collect diverse moving trajectories in the real world for training. Previous works simplify the setting by manually designing different trajectories in the simulator~\cite{akinola2021dynamic}, \emph{e.g.,} line, sine, or circle. However, the bias in trajectories will degrade the generalization in unseen environments.

To this end, inspired by AD-VAT~\cite{zhong2018ad}, we introduce an adversarial reinforcement learning framework to generate trajectories for learning robust dynamic grasping policy automatically, referred to as GraspARL.

We consider the robot arm and object mover as two solo agents, playing a min-max game during training, which produces an auto-curriculum learning process. 
Especially at the very beginning, the object trajectory is nearly random.
As the training proceeds, the mover automatically learns more diverse and complex trajectories to move the object to get away from being picked up. Meanwhile, the robotic arm learns to approach and grasp the object at the beginning naively, then gradually improves its tracking and grasping ability jointly with the moving strategy.

As the game is asymmetrical, it remains challenging to train the two policies with a zero-sum game.
To cope with such problems, we further propose an object-geometry-aware reward (OGAR).
OGAR is a reward function that facilitates the robot to learn dynamic grasping by indicating the potential grasping area, meanwhile, our OGAR also considers penalty for collision with object during grasping.

The contributions of our work can be summarized as follows:
1) We propose an adversarial RL framework to improve the generalization of dynamic grasping in unseen trajectories.
To the best of our knowledge, this is the first adversarial reinforcement learning framework for robust dynamic grasping.
2) We propose an object-geometry-aware reward (OGAR) to make the adversarial training process more stable and the grasping policy safer.
3) We comprehensively evaluate and analyze the effectiveness of the framework in the simulator, and successfully deploy the learned policy on the real-world robot.

\section{RELATED WORK}

\subsection{Dynamic Grasping}

With the advances in deep learning~\cite{lecun2015deep}, we have witnessed significantly improvement in static object grasping~\cite{morrison2018closing,kalashnikov2018scalable,mousavian20196,fang2020graspnet}. Some methods also show the potential of handling slightly moving objects, \emph{e.g.}, replacing objects once during grasping~\cite{morrison2018closing,viereck2017learning,song2020grasping}, or pushing objects with a constant force at beginning~\cite{hu2020reaching}. However, it remains infeasible to handle continuous moving objects by these methods.
It is tedious to develop a system for dynamic object grasping, which requires real-time visual object tracking~\cite{issac2016depth,tremblay2018deep,wang2019densefusion}, trajectory modeling and prediction~\cite{zhang2014fast,farahi2020probabilistic}, and reactive motion control~\cite{islam2021provably,marturi2019dynamic,schmitt2019planning}. 

Previous works on dynamic grasping focus on developing a specific module, especially for visual tracking~\cite{allen1993automated,husain2014realtime,zhao2017visual}, and object grasping~\cite{islam2021provably,migimatsu2020object,ye2018velocity,menon2014motion,de2021dual,suzuki2015grasping,mirrazavi2017dynamical} to realize dynamic grasping.
However, most of these works can only handle the case that the trajectories of objects are with simple patterns, such as line~\cite{islam2021provably,ye2018velocity,menon2014motion,de2021dual,suzuki2015grasping}, sine~\cite{ye2018velocity}, arc~\cite{ye2018velocity}, circle~\cite{allen1993automated}, and quadratic curve~\cite{kim2014catching}.
Recently, Akinola, Iretiayo, \emph{et al}.~\cite{akinola2021dynamic} introduce a complete solution with reachability and motion awareness for dynamic grasping. The motion-awarness includes LSTM for better generalization on line, sine and circle. However, it is of poor generalization in unseen and complex trajectories, \emph{e.g.}, random movement.
In this paper, we introduce an adversarial reinforcement learning solution that automatically generates diverse moving trajectories of objects in training, to further improve the generalization of the grasping policy in case of different moving trajectories.

It is also attractive to train a reactive policy via reinforcement learning for dynamic grasping. A recent work~\cite{wang2020multi} employs multi-task reinforcement learning to dynamic grasp cubes with a mobile robot. 
Our setting differs from~\cite{wang2020multi} in that the base of our robot is fixed, constraining the reachable area for the robot.
DHER~\cite{fang2019dher} also related to our task, a dynamic hindsight experience replay buffer is proposed to handle dynamic goals. However, only dynamic reaching is achieved in this method.

\subsection{Adversarial Reinforcement Learning}
It is not a new concept to improve the robustness of the policy network via adversarial reinforcement learning~\cite{pinto2017robust,huang2017adversarial,mandlekar2017adversarially}.
In the simplest setting, the adversary is not a physical player, only add disturbance at the action and observation of the policy.
In self-play~\cite{bansal2018emergent,silver2017mastering}, both protagonist and adversary are physical players, complex skills can be learned via competition. 
There are also some tasks that are asymmetric, and self-play also shows the power to help the protagonist to learn better policy. Zhong \etal \cite{zhong2018ad} proposed a method for one agent to track another agent to form an adversarial game to learn a more robust tracker in dynamic object tracking. Tang \etal ~\cite{tang2020learning} adopted a self-play chasing game for learning an effective locomotion controller for agile locomotion. 
Differently, these works only consider following a moving object, grasping object is more difficult as more accurate manipulation policy is required. 
Thus, we propose the object-geometry-aware reward (OGAR) to encourage the robot to learn a safe yet efficient grasping policy.

\section{METHOD}
In this section, we first introduce the formulation of the adversarial reinforcement learning framework for dynamic grasping. 
Next, we describe the details of the reward structure. Then, we further illustrate the state representation for both agents (robotic arm and mover). At last, we introduce the training strategy for the framework.

\subsection{Move-and-Grasp Game}
In this work, we formulate the ``dynamic grasping'' problem as a Two-agent Markov Game, namely ``Move-and-Grasp" Game.
In this game, the robot (Noted as agent $1$) aims to control the movement of the robotic arm to grasp a target object, which is continuously moved by an adversarial mover (Noted as agent $2$) to escape the grasping.    
In formal, the game is govern by a tuple $<\mathcal{S}$, $\mathcal{A}_1$, $\mathcal{A}_2$, $\mathcal{O}_1$, $\mathcal{O}_2$, $\mathcal{R}_1$, $\mathcal{R}_2$, $\mathcal{P}$, $\gamma_{1}$,  $\gamma_{2}>$, where $\mathcal{S}$, $\mathcal{A}$, $\mathcal{O}$, $\mathcal{R}$, $\mathcal{P}$ and $\gamma$ represent the state space, action space, observation space, reward function, the state transition probability and discount factor.
For every timestep $t$, the robot and mover receive observation $o_{1, t}$ and $o_{2,t}$, and take action $a_{1,t}\sim\pi_1(o_{1,t})$, $a_{2,t}\sim\pi_2(o_{2,t})$ simultaneously, where $\pi$ represents the policy. After action executed, robot and object will receive reward $r_{1,t}(s_t,a_{1,t},a_{2,t})$ and $r_{2,t}(s_t,a_{1,t},a_{2,t})$, along with next state $s_{t+1}$ updated according to state transition probability $\mathcal{P}$. 
The goal of each agent $i$ is to maximum its expected total reward
\begin{equation}
    \mathbb{E}_{a_{i, t}\sim\pi_1(o_{i, t})}[\sum_{t=1}^{T}\gamma_i^{t-1}r_{i, t}(s_t,a_{1, t},a_{2,t})]
\end{equation}
Intuitively, once the $r_1 + r_2 = 0$, the two agents will play a min-max game, \ie, the gain of one agent will cause the loss of the other.

\begin{figure*}[t]
  \centering
  \includegraphics[trim=10 100 25 100,clip,width=\linewidth]{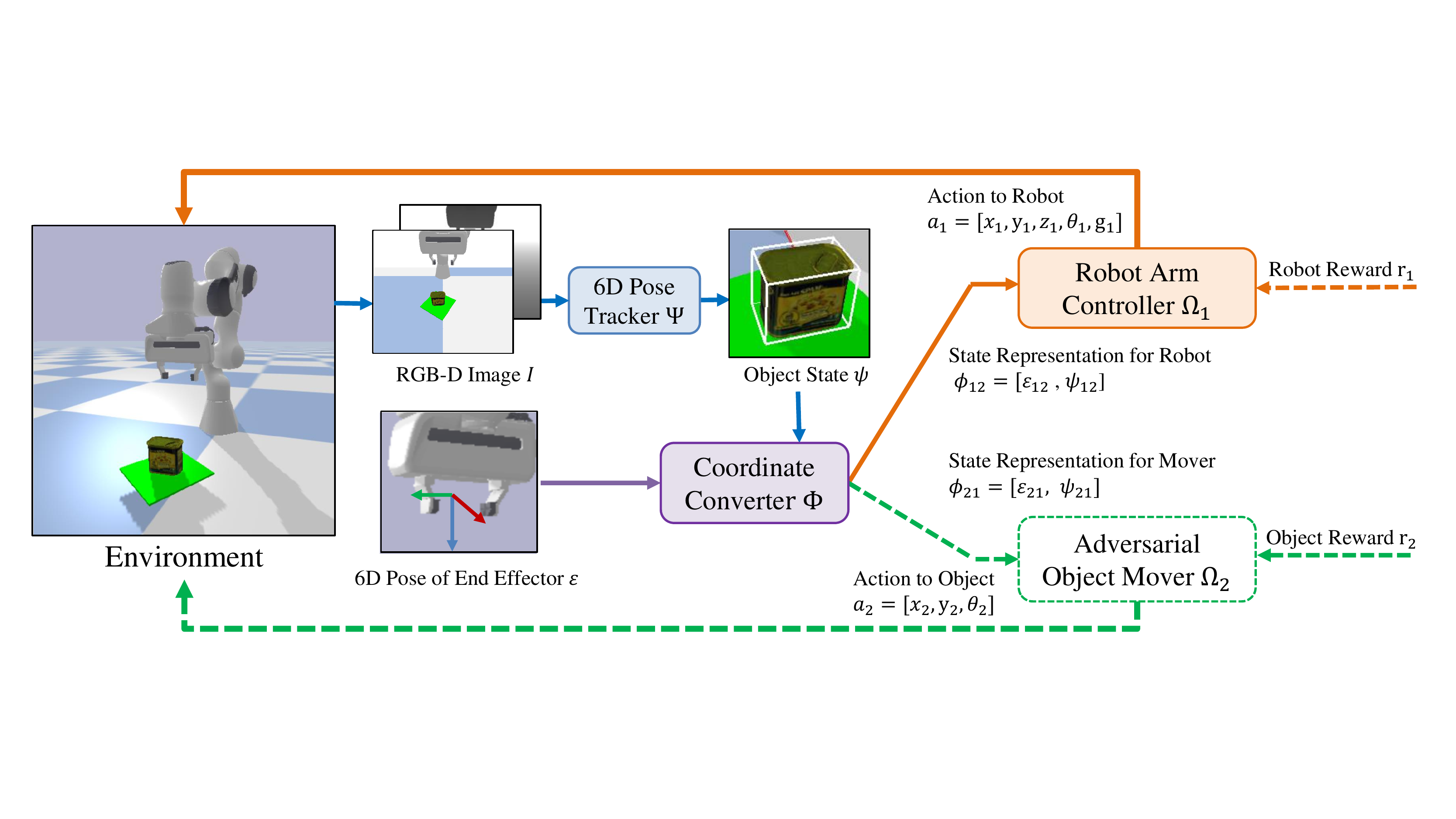}
  \caption{\textbf{The Adversarial Reinforcement Learning Framework for Dynamic Grasping.}
  For perception, a 6D pose tracker receives the RGB-D image and output 3D bounding box of the target object to the coordinate converter.
  The coordinate converter transform both the robot and object state to the agent-centric coordinate system.
  During training, the robot arm controller and object mover are viewed as opponents. The arm controller learns how to grasp the object, while the object mover learns to generate diverse moving patterns to avoid the object being grasped.
  }
  \label{fig:framework}
\end{figure*}

\subsection{Reward for Adversarial Dynamic Grasping} 
The typical adversarial learning setting for two agents is a zero-sum game. However, this formulation is not suitable for our asymmetric game, the goal for the robot is to success grasp the object, the goal of the object is to make the grasp fail, which is much easier than the robot's goal since the object's action space is smaller than robot's action space. 
To address the problem, we propose an asymmetric reward structure to stabilize adversarial learning. 
\subsubsection{Reward for robot}
For the robot to learn dynamic grasping, the first step is to approach the object. This can be effectively achieved by setting the distance $d_{12}$ between robot and object as an initial penalty. More specifically, we define distance $d_{12}$ as follow:
\begin{equation}
d_{12}=\sqrt{(g_x-o_x)^2+(g_y-o_y)^2+(g_z-o_{bz})^2}
\end{equation}
where $g_x, g_y, g_z$ represent for the x, y, z coordinate of the gripper center point, $o_x, o_y$ represent for the x, y coordinate of the object center point, $o_{bz}$ represent for the max z value of the object bounding box, all coordinates are with respect to the world coordinate frame.

However, only applying distance $d_{12}$ as reward is easy for the robot to collide with the object, since different object has different geometry. Thus the goal of the robot is more challenging than the goal of the object, which makes the dynamic grasping game asymmetric. To address the issue, when the center point of the gripper is moving inside the object bounding box, a positive reward $R_{b}$ is set to encourage the robot to grasp the object. Such reward can make robot aware the geometry of the object which also indicates the potential grasping area for the robot. In real scene, a robot collides with an object might cause damage to the object. Thus, we add a `collision check' between the object and robot. Unless the collision is happened due to success grasp, a penalty $P_{coll}$ will set to encourage robot to less collide with the object.

For each step the robot takes, we add a small step penalty $P_{time}$ to encourage the robot to grasp the object with less steps. Since the object has real dynamics, once the object falls out of the plate, we consider the robot failed to grasp the object, $r_{1}$ will set to $P_{out}$ and terminate the episode. Grasping is considered success if the object's height is a certain distance higher than the initial height after the robot lifts the gripper, then $r_{1}$ will set to $R_{s}$ and terminate the episode. 

We name our reward object-geometry-aware reward (OGAR), The total robot reward is as follow:
\begin{equation}
r_{1}=\left\{
\begin{array}{lll}\vspace{1ex}
R_s       &     {grasped}\\ \vspace{1ex}
P_{out}     &     {out\ of\ plate}\\ \vspace{1ex}
\overline{\mathbbm{1}^{b}}P_{dis}+\mathbbm{1}^{b}R_{b}+P_{coll}+P_{time}     &     {others}
\end{array} \right.
\label{reward_robot}
\end{equation}

 where $R_{s}>0$, $P_{out}<0$, $P_{dis}<0$, $R_{b}>0$, $P_{coll}<0$, $P_{time}<0$, $\mathbbm{1}^{b}=1$ when the grasping point is inside the bounding box of the object.

\subsubsection{Reward for mover}
To avoid the object being grasped, the mover should keep away with the gripper. 
Thus, $P_{dis}$ is also used as a reward for the mover keeping a considerable distance from the gripper. Additionally, we also set a safety distance $d_{safe}$ for the object. When the distance $d_{12}$ is smaller than safety distance, a penalty $P_{close}$ is given to the mover to indicate that the gripper is too close. We did not use bounding box to indicate the safety distance since when the robot approaches the bounding box of the object, the distance between the robot and object is already too close. The total reward for the object is as follow,
\begin{equation}
r_{2}=-P_{dis}+\mathbbm{1}^{safe}P_{close}
\end{equation}
where $P_{dis}<0$, $P_{close}=-d_{safe}$, $\mathbbm{1}^{safe}$=1 when $d_{12}<d_{safe}$.

\subsection{State Representation}
For each step, 6D pose tracker ($\Psi$) first get RGB-D image ($I$) from the environment and estimates the 6D pose of the object, then inference the orientated bounding box of the object with the corresponding object mesh, and we use the key points of the orientated bounding box to represent object state ($\psi$). 
For robot state, we use the 6D pose of end effector ($\varepsilon$), which can directly get from the robot in simulated or real-world environments.

The relative relation is the key information that the robot and object need to know during dynamic grasping. Thus, we adopt a coordinate converter ($\Phi$) to convert both object state and 6D pose of end effector ($\varepsilon$) to relative state.
For robot, the 6D pose of end effector ($\varepsilon$) and object state ($\psi$) transform with respect to robot base coordinate frame and end effector coordinate frame respectively. The transformed state representation for robot is $\phi_{12}=[\varepsilon_{12},\psi_{12}]$, which input to robot arm controller ($\Omega_1$). Then the robot arm controller ($\Omega_1$) takes action $a_1=[x_1,y_1,z_1,\theta_1,g_1]$, where $x_1,y_1,z_1$ represents the 3D position relative to the object, $\theta_1$ represents the yaw orientation relative to the object, $g_1$ represents the gripper open/close command.
For object, the 6D pose of end effector ($\varepsilon$) and object state ($\psi$) both transform with respect to object base coordinate frame. The transformed state representation for object is $\phi_{21}=[\varepsilon_{21},\psi_{21}]$, which input to adversarial object mover ($\Omega_2$). Then the adversarial object mover takes action $a_2=[x_2,y_2,\theta_2]$, where $x_2,y_2$ represents the 2D translation, $\theta_1$ represents the yaw rotation.
Utilizing the relative pose can help the robot generalize to the scenario that the robot and object's absolute pose is different from training.

\subsection{Training Strategy}
 To encourage the robot to learn to reach the object at the early stage, we first set $\gamma_1
$ to a small value. Then we incrementally increase $\gamma_1$ to guide the robot to learn grasping policy from a longer-term perspective.

We simultaneously train the robot and mover from scratch. However, such training process will cause the robot to overfit to concurrent object policy. Thus, we create a model pool to save different object policies during training. After converging, we select the robot policy with the best performance and finetune the robot policy with the object policy randomly sampled from the model pool. To generate more diverse trajectories for finetuning, we randomly stack the state given to the object, and the action taken by the object.
For adversarial training phase, we use the grounded state as the input of policies.

\begin{figure*}[t]
  \centering
  \includegraphics[trim=0 320 0 0,clip,width=\linewidth, scale=1.1]{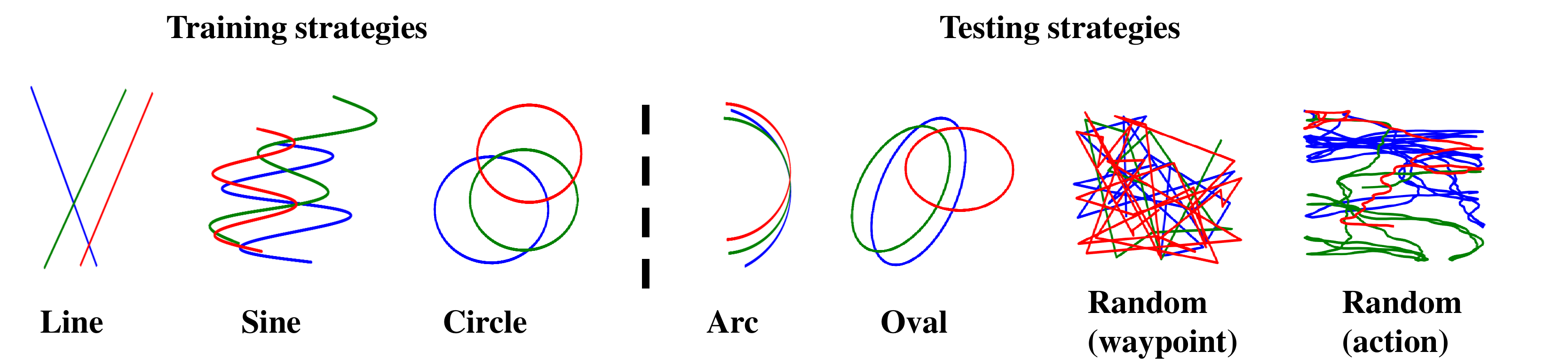}
  \caption{\textbf{Top-view of Different Motion Patterns of the Target Object.}
  Different colors indicate different episodes.
  The left three (Line, Sine, and Circle) are used for training baselines. The right four are for the testing. Note that our adversarial RL method does not use any pre-defined trajectories for training.}
  \label{fig:strategies}
\end{figure*}

\section{EXPERIMENTS}
\subsection{Experimental Setup}
    \subsubsection{Simulation Environment}
    We use Pybullet~\cite{coumans2016pybullet} simulator to build our simulation environments. We chose the Franka Panda robot as our robot agent and set it at the origin of the world. Similar to the setting in ~\cite{akinola2021dynamic}, there is a plate placed with a target object in the center. The object can be pushed out of the plate and fall to the ground.
    The plate can move along the $x$ and $y$ axis, and controlled by the adversarial mover. 
    To guarantee the robot action has feasible inverse kinematics, the robot moving area insides the $x$ axis varies from $0.2m$ to $0.6m$, the $y$ axis varies from $-0.5m$ to $0.5m$, and the $z$ axis varies from $0.055$ to $0.3m$ with respect to the world coordinate. 
    To guarantee the object will not move beyond the robot moving area, the object moving area insides the $y$ axis varies from $-0.45m$ to $0.45m$, and the $x$ axis varies from $0.25m$ to $0.55m$ with respect to the world coordinate.
    
    We define seven motion patterns for simulating different object moving trajectories while training and testing.
    Fig.~\ref{fig:strategies} shows the typical examples of the motion patterns.
    All baseline methods are trained on the left three moving strategies: \textit{Line}, \textit{Sine}, and \textit{Circle}.
    Note that our method does not use these pre-defined motion patterns in the training phase, as the adversarial mover can automatically generate the moving trajectories.
    For testing, all methods are evaluated on a set of unseen patterns, including two pre-defined trajectories (\textit{Arc} and \textit{Oval}) and two random moving strategies. 
    The random moving strategies are more complex, including random sampling waypoint (denotes as `random (waypoint)') and randomly adjusting angular velocity (denotes as `random (action)').
    To diversify the trajectories, we randomize the parameters of each pattern at the beginning of the episode. 
    For example, we generate the \textit{sine} by randomly configuring the starting point, amplitude, angular velocity, and rotation angle of the \textit{sine} function.
    For rotation motion, we set delta rotation to a constant value in the training phase, while in the testing phase, the delta rotation for each step is sampled from a Gaussian distribution.

\begin{figure}[t]
  \centering
\includegraphics[trim=55 80 260 50,clip, scale=0.4]{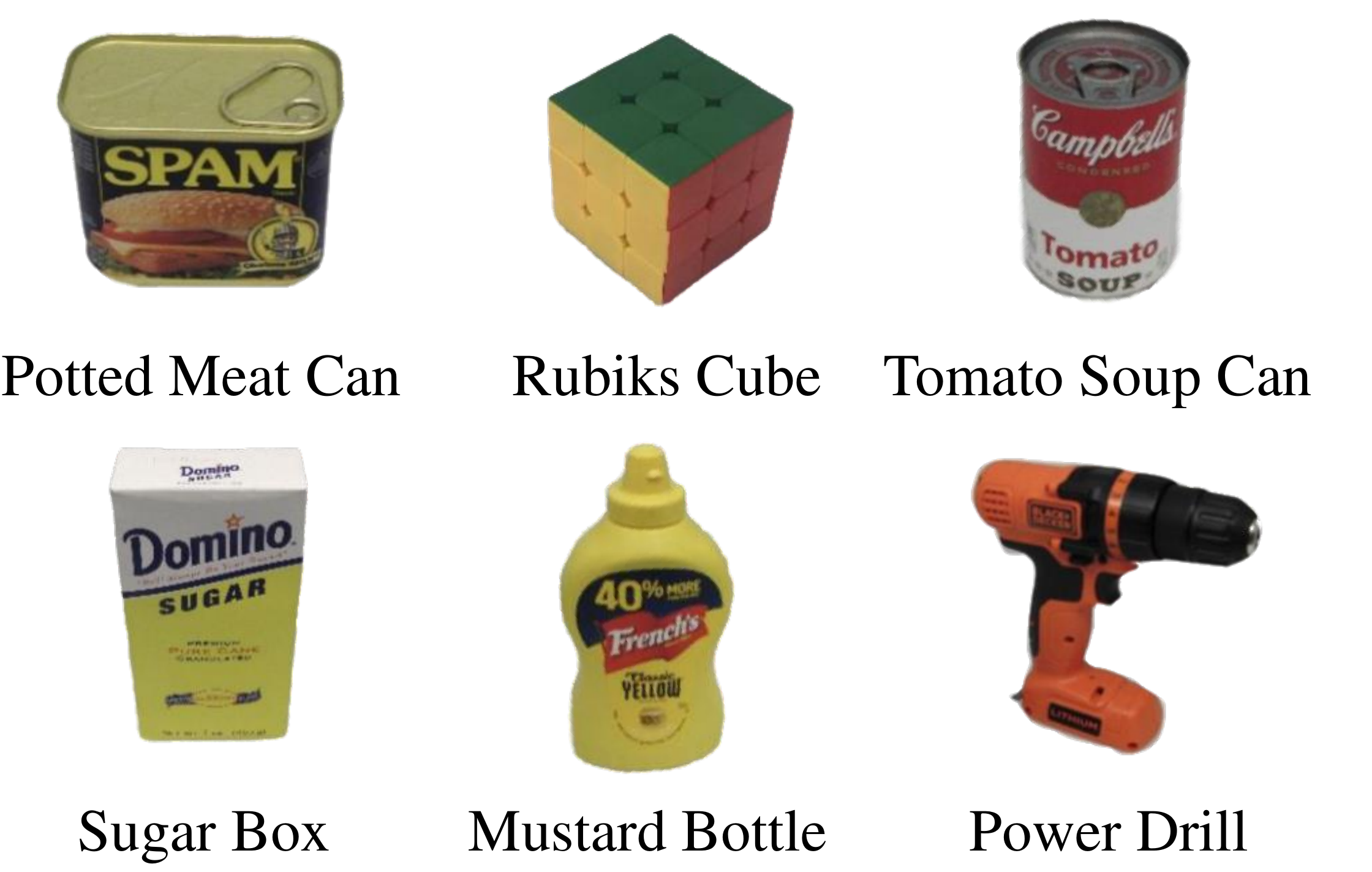}
  \caption{\textbf{Selected Objects from YCB Dataset~\cite{calli2015ycb}}}
  \label{fig:object}
\end{figure}

We choose six objects with different sizes and shape from YCB dataset~\cite{calli2015ycb}, as shown in Fig.~\ref{fig:object}. Specifically, the selected objects are potted meat can, rubiks cube, tomato soup can, sugar box, mustard bottle, and power drill.
Focusing on evaluating the motion generalization, they all are used for training and testing. 
    
    \subsubsection{Baseline}
    We choose two learning-based methods and one non-learning based methods as our baselines. During training, the state is the grounded state directly get from the simulation for both baselines and our method, such as the position and orientation of robot and object. 
    
    \begin{itemize}
    \item Unscented Kalman Filter with Control Policy (UKF+CP)~\cite{julier1997new}, is more suitable for non-linear trajectories to predict object trajectory and rotation. We use PID as our control policy, which takes the 6D pose of the object as the policy input, and grasping the object with predefined grasp poses.
    
    \item Vanilla Reinforcement Learning (RL) is optimized by A3C~\cite{mnih2016asynchronous} to learn grasp different moving objects. The reward for it is \begin{equation}
    r'_1=\left\{
    \begin{array}{lll}
    R_s       &     {grasped}\\
    P_{out}     &     {out\ of\ plate}\\
    P_{dis}    &     {others}
    \end{array} \right.
    \label{con:rlreward}
    \end{equation} 
    where $P_{dis}$, $P_{out}$, $R_s$ are the same as ~\ref{reward_robot}.
    \item Dynamic Grasping with Reachability and Motion Awareness (DGRMA)~\cite{akinola2021dynamic} uses reachability and motion-aware grasp planning module, adaptive motion generation module and recurrent neural network(RNN) module to improve the grasping performance. We use their core modules and adapt them to our environment as a baseline.  
    \end{itemize}
    
\subsubsection{Implementation Details}
We introduce the details of the policy network, object pose tracker and the hyper parameters as followings:

\textit{Object Pose Tracker}:
We choose an RGB-D based 6D pose estimator DenseFusion~\cite{wang2019densefusion} to estimate 6D pose of object. DenseFusion fusions the depth and RGB image to estimate the object 6D pose and further refine with an iterative pose refinement module. We recollect the data to retrain the model from scratch for our simulation experiments.

\textit{Policy Network}: For both the robot and mover, we follow the same Conv-LSTM network architecture as AD-VAT~\cite{zhong2018ad}, but with different input and output.

\textit{Hyper Parameters}:
For the reward of the robot, we try three different values for time penalty $P_{time}$ including -0.0005, -0.005 and -0.05. The result shows that -0.05 is too large compared to our distance reward $R_{dis}$, and -0.0005 is too small to indicate time cost. We try three different values for bounding box reward $R_b$ including 0.05, 0.1 and 0.2. The $P_{coll}$ is sett to $-R_b$ to indicate robot that collides with the object inside object's bounding box area is risk, the result shows that $R_b=0.1$ and $P_{coll}=-0.1$ can help robot to learn high success grasp with less collision. For the out-of-plate penalty $P_{out}$, we try three different values, including -1, -0.1 and -0.05, the result shows that -1 penalty will cause the robot sometimes to avoid reaching the object, -0.05 penalty will not let the robot learn to avoid object collison. The $R_{s}$ is set to 10 to guarantee the discount success reward in steps before success can be larger than other penalty rewards. For the reward of object, we try three different values for $d_{safe}$ including 0.1, 0.2 and 0.3, we find when $d_{safe}=0.1$, robot is already very close to object, and for $d_{safe}=0.3$ and $R_{dis}$, it is hard to indicate the distance between two agents for the object.

For the adversarial learning rate of the robot and object, we tried three different learning rates, including 0.0001, 0.001 and 0.01, both agents can converge fast under learning rate of 0.001. For curriculum $\gamma_1$, the final $\gamma_1$ is set to 0.96 to encourage the robot considers more steps. We try four different initial $\gamma_1$ values, including 0.2, 0.5, 0.8 and 0.9, and three different $\gamma_1$ increase coefficients, including 5e-7, 5e-6 and 5e-5, the robot can achieve high performance and converge fastest under initial $\gamma_1$ of 0.5 and the $\gamma_1$ increase coefficient of 5e-6. When finetuning the robot, we decrease the learning rate of the robot to 0.0001 for learning a more accurate policy. For training, finetuning, and testing, we set the velocity of movement for the robot to 0.15 m/s, and the velocity of rotation for the robot is 0.75 rad/s.

\subsubsection{Evaluation Metrics}
Two metrics are defined for quantitative comparison. 
\begin{itemize}
\item Success Rate (SR) is the ratio of success grasp number to total test episode number. The higher SR indicates better performance.
\item Average Episode Length (AEL): SR cannot fully measure the performance since the result with the same SR may use different time steps. So we also define AEL to be the average number of steps that the robot takes in each testing episode. If the object falls from the mover, we consider the grasping failed and set the AEL to be the maximum steps of the episode. The smaller AEL indicates better performance.
\end{itemize}

\begin{figure*}[t]
  \centering
  \includegraphics[trim=10 130 60 0,clip,width=\linewidth,
  ]{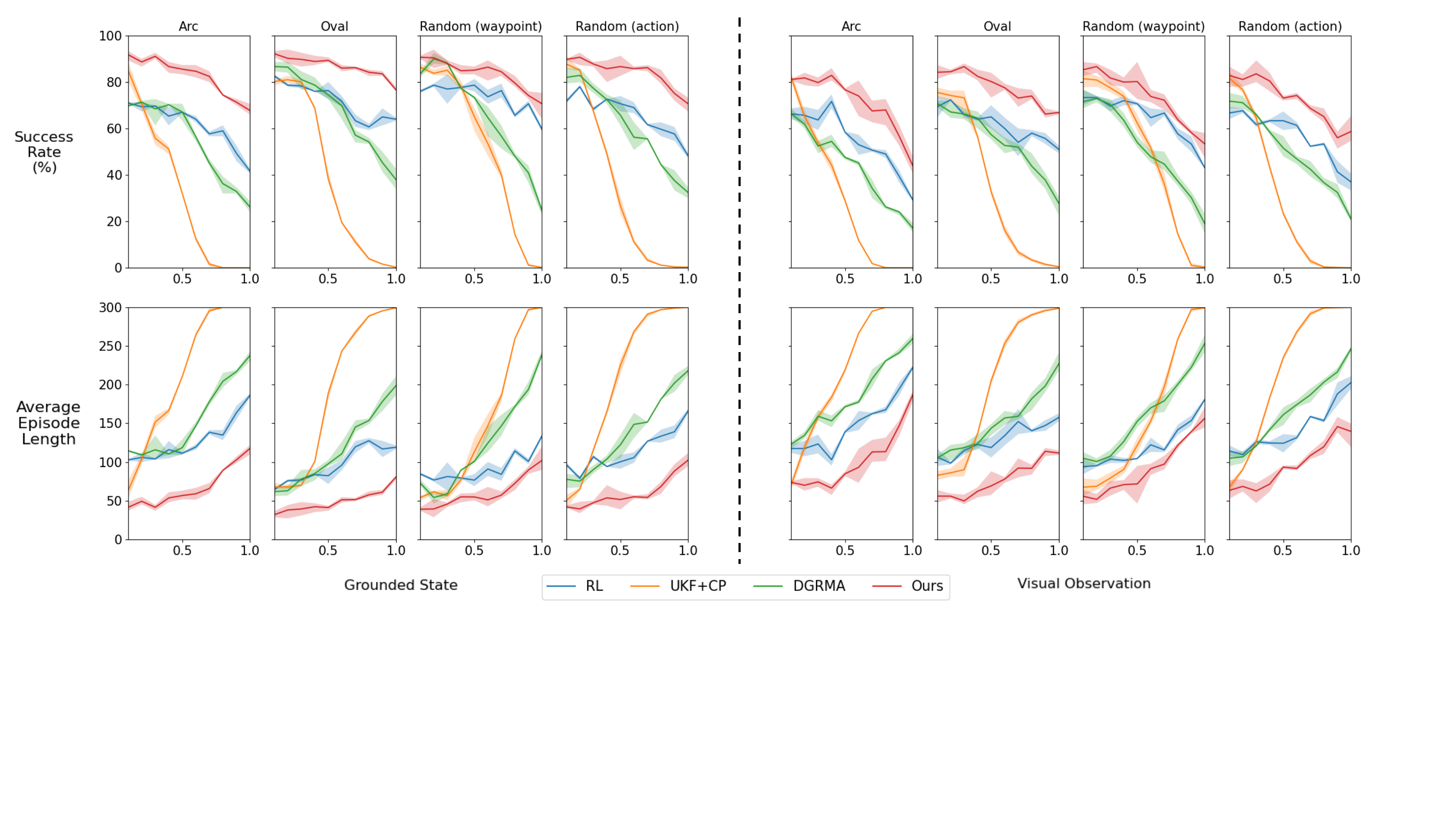}
  \caption{\textbf{Quantitative Comparison}. 
  Each sub-figure illustrates the performance of four different methods under each motion pattern. The horizontal axis is the speed ratio of the mover velocity with respect to the robot velocity.
  The vertical axis of the top rows represents the success rate, the higher the better.
  The vertical axis of the bottom rows represents the average episode length, the lower the better.
  The left four columns are results using the grounded state as the policy input, the right four columns are results using the visual observation as the policy input.
  Note that, those moving strategies are unseen for all methods in the training phase.
  }
  \label{fig:baseline}
\end{figure*}

\subsection{Quantitative Comparison}
\label{sec:baseline}
To fully show the effectiveness of our method, we conducted the experiment both using the grounded state and visual observation. For mover velocity setting, we set the ratio of the mover's velocity with respect to the robot's velocity from 0.1 to 1.0. More specifically, we divided the action bound space (\emph{i.e.,} speed space) into ten different intervals from low to high speed with interval of 0.1. For each method, we choose three random seeds. For each random seed, we will test 50 episodes (an episode will run at most 300 steps) in simulator for each test trajectory, speed interval and object. For each speed interval, the speed of the mover is randomly sampled according to the min and max bounds, the final result is averaged across different random seeds and objects. The result is shown in Fig.~\ref{fig:baseline}.

The result shows that UKF+CP achieves high success rate when the velocity ratio is low since any object trajectories can be considered a straight line at such speed interval. However, as the velocity ratio increases, the UKF prediction error increases, the control policy is also hard to adjust the pose of the robot quickly, resulting in the success rate dramatically decreasing at high speed. DGRMA also achieves high success rate for speed ratio lower than around 0.4, especially for oval and multipoint trajectories. DGRMA also has a more robust performance at high speed than UKF+CP. However, it is not robust enough compared with the RL and our method. This is because reinforcement learning learns the policy by trial-and-error instead of supervised learning. Thus, the RL can learn the essential difference and generalize better for unseen trajectories. Although RL is robust across all speed ratios compared with other baselines, the performance is still lower than our method, since RL is only trained with line, sin, and circle, limiting the generalization for unseen trajectories. Our method outperforms all the baselines in all unseen trajectories and all speed ratios for both SR and AEL. Besides, our method is robust for all different trajectories, especially for the arc. Other baselines all have the worst performance for the arc, but our method still has high SR, which fully shows the robustness of our method. 

For visual observation, two problems might occur: incorrect detection of symmetric objects and occlusion between the object and others. Both problems will affect the estimated 6D pose and further increase difficulty for grasping. Thus, as shown in Fig.~\ref{fig:baseline}, there are performance drops for most of the methods. DGRMA chooses the grasp pose according to object 6D pose. Thus, when the detection is incorrect, the grasp pose will get affected, such as for rubiks cube, we observe that when using visual observation, the success rate of rubiks cube dramatically decreases to nearly zero. Besides, DGRMA also has problem when getting occlude with the object during grasping. Thus, the performance of DGRMA is downgraded obviously. For RL and our method also drops under visual observation, since they face the same problem as DGRMA.
UKF+CP only drops a little under visual observation, this is because the grasping policy we implement for this baseline is first tracking while keeping a certain distance along the z axis of the object, then performing a top-down grasping policy. Thus, the occlusion problem hardly occurs during the grasping process. Besides, since UKF uses a top-down grasping policy, most of the objects in our dataset are smaller than the width of the gripper, thus, even the estimated orientation is incorrect, it won't affect the grasping. However, our method still has the best performance compared to all other baselines, even with drops in performance, which shows the robustness of our dynamic grasping system.

\begin{figure}[]
  \includegraphics[trim=10 100 530 0,clip,width=\linewidth 
  ]{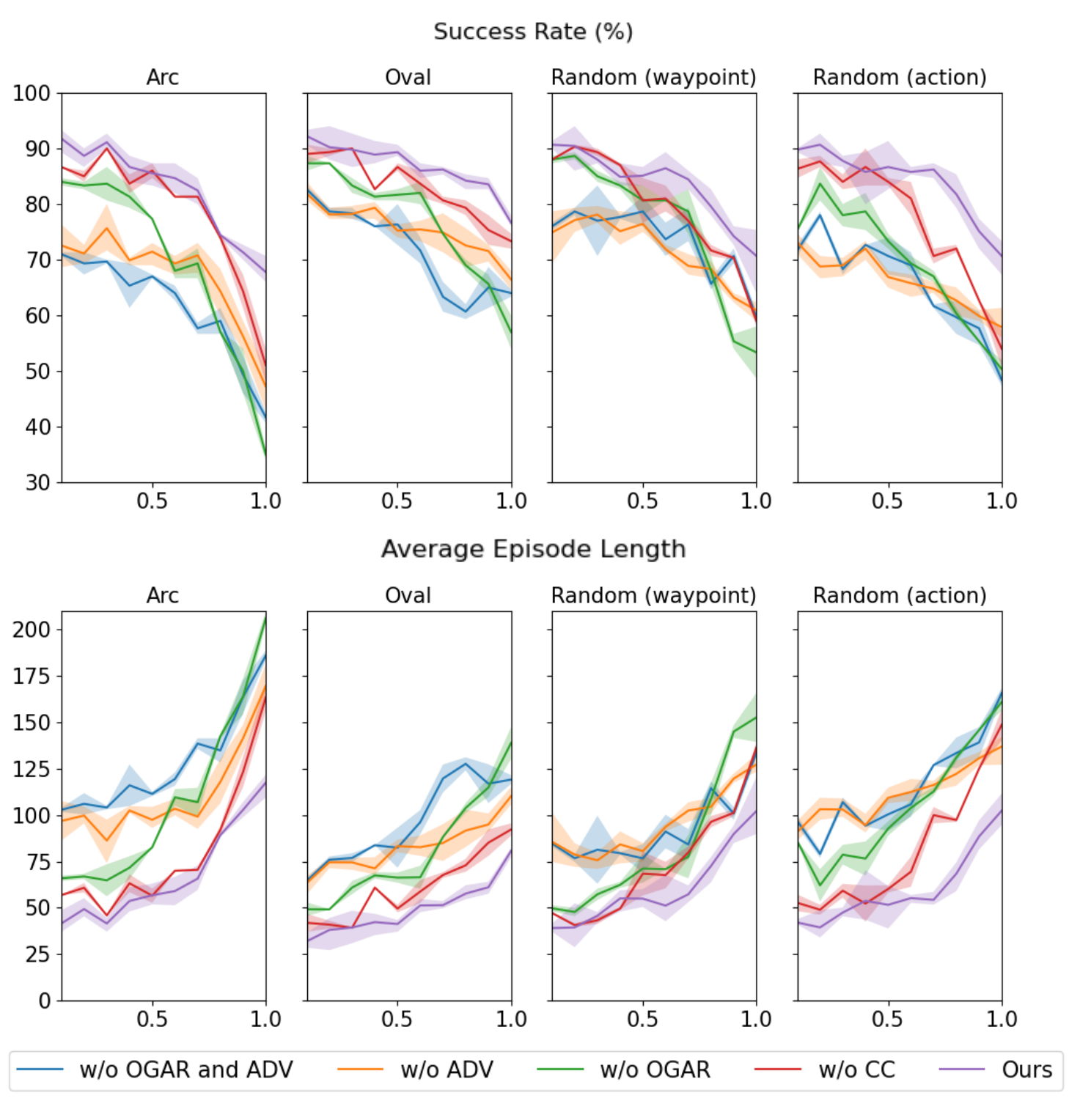}
  \caption{\textbf{Ablation Study.}
  Each sub-figure illustrates the performance of our method without different module under the same moving strategy. 
  The horizontal axis represents the speed ratio. The vertical axis of top rows are success rate, the higher the better.
  The vertical axis of bottom rows represents the average episode length, the smaller the better.
  }
  \label{fig:ablation}
\end{figure}

\subsection{Ablation Study}
 To show the necessity of different components of our method, we conduct an ablation study by comparing our method with: 
 1) our method without adversarial learning (w/o ADV), 
  2) our method without the object-geometry-aware reward (w/o OGAR). For this ablation, the reward for the robot is replaced by Eq.~\ref{con:rlreward}, And we apply zero-sum reward for the object.
  3) our method without the collision check reward (w/o CC).
  4) our method without ADV and OGAR (\emph{i.e.}, the naive RL baseline).
The experiment setting follows the same setting in Sec.~\ref{sec:baseline}, except only the grounded state is used in the ablation study.

 As shown in Fig.~\ref{fig:ablation}, when only considering a single component, our method without ADV has the worst performance, which shows that adversarial reinforcement learning contributes most to our method. Our method without OGAR and ADV has the lowest SR and highest AEL. However, compared to our method without ADV, the performance does not decrease much, which further indicates the significance of ADV. For our method without OGAR reward, the success rate is also decreased and reduced more at the high speed ratio, which shows that OGAR reward does help to improve the effectiveness of the asymmetric adversarial training.
 Although our method without collision check still has high SR and low AEL, the performance still degrades compared with our method, which shows that collision check does help the robot learn better policy.

\subsection{Real-World Deployment}
To show the transferability of our method in the real-world scenario,
we further deploy our dynamic grasper on a Franka Panda robot.
To be specific, we use an Arduino programmable car as the mover to carry the target objects. 
The car random moving on the desk, and the velocity of the car is around $5 cm/s$.
We chose two objects as the target, the potted meat can and the mustard bottle.
As shown in Fig.~\ref{fig:teaser}, the base of the robotic arm is fixed on one side of the table, and the RGB-D camera is mounted on the camera crane on the other side of the desk. 
At the beginning of each testing episode, we put the object on the top of the car and release the car on one side of the table. 
Then, we perform the trained model on a desktop computer with 8 Core Interl CPU (i7-7700K) and 1 NVIDIA GPU (Titan Xp). 
The episode is considered a success if the robot can pick up the object before the car reaches the other side of the table.

Following the experiment in~\cite{akinola2021dynamic}, we repeat the grasping experiment 5 times for each object. The success rate of potted meat can is $4/5$, and mustard bottle is $3/5$. Since our robot performs a top-down grasp policy, the robot needs to accurately reach the cap of mustard to catch the mustard bottle. Thus, it is harder for the robot to grasp the mustard bottle than meat can. Please refer to the supplementary videos for more vivid examples.

\section{CONCLUSION}
This study is the first attempt to formulate the dynamic grasping problem as a ``move-and-grasp" game and use adversarial RL to train the grasping policy and object moving strategies jointly.
Compared with the baseline, our method does not require any human pre-defined motion patterns for moving objects during training.
Nevertheless, quantitative results on the simulator show 
that our method significantly outperforms the baselines on various unseen motion patterns with different speeds configuration.

For future work, we will pay attention to leveraging better 3D perception methods so as to further improve the dynamic grasping on objects with complex shapes.
Besides, we believe the adversarial learning mechanism could benefit many other robotic tasks that have many possible moving strategies, we will apply our method to other robot-to-object or robot-to-robot interaction tasks.






\section*{ACKNOWLEDGEMENT}
This project was supported by National Natural Science Foundation of China — Youth Science Fund (No.62006006).
We would also like to thanks the lab mates
for the helpful discussion.




\bibliographystyle{IEEEtran}
\bibliography{IEEEabrv,ref}

\end{document}